\documentclass{article}

\usepackage{PRIMEarxiv}

\usepackage[utf8]{inputenc} 
\usepackage[T1]{fontenc}    
\usepackage{hyperref}       
\usepackage{url}            
\usepackage{booktabs}       
\usepackage{amsfonts}       
\usepackage{nicefrac}       
\usepackage{microtype}      
\usepackage{lipsum}
\usepackage{fancyhdr}       
\usepackage{graphicx}       
\graphicspath{{media/}}     

\usepackage{siunitx}
\usepackage{longtable}
\usepackage{subcaption}
\usepackage{multirow}

\usepackage{newfloat}
    \DeclareFloatingEnvironment[name={\textbf{Supplementary Table}},fileext=lst,listname={List of 
Supplementary Tables}]{supptable}

\usepackage{newfloat}
    \DeclareFloatingEnvironment[name={\textbf{Supplementary Figure}},fileext=lst,listname={List of 
Supplementary Figure}]{suppfig}

\title{A Position Statement on Endovascular Models and Effectiveness Metrics for Mechanical Thrombectomy Navigation, on behalf of the Stakeholder Taskforce for AI-assisted Robotic Thrombectomy (START)
\thanks{\textit{\underline{Citation}}: 
\textbf{Robertshaw H, Barnes A, Blakelock P, Blanc R, Crossley R, Fahrig R, Hassan AE, Jackson B, Karstensen L, Kaur N, Kowarschik M, Lynch J, Mathis-Ullrich F, Meglan D, Mendes Pereira V, Ourak M, Pantano M, Sadati SMH, Taylor-Gee A, Vercauteren T, White P, Granados A, Booth TC. A position statement on endovascular models and effectiveness metrics for mechanical thrombectomy navigation, on behalf of the stakeholder taskforce for artificial intelligence–assisted robotic thrombectomy (START). \textit{Journal of the American Heart Association}. 2026;15:e044931. DOI:\url{https://doi.org/10.1161/JAHA.125.044931}}} 
}

\small
\author{\normalfont
Harry Robertshaw\textsuperscript{1},
Anna Barnes\textsuperscript{2},
Phil Blakelock\textsuperscript{3},
Raphael Blanc\textsuperscript{4},
Robert Crossley\textsuperscript{5},
Rebecca Fahrig\textsuperscript{6,7}, \\
Ameer E Hassan\textsuperscript{8,9},
Benjamin Jackson\textsuperscript{1},
Lennart Karstensen\textsuperscript{10},
Neelam Kaur\textsuperscript{11},
Markus Kowarschik\textsuperscript{6},\\
Jeremy Lynch\textsuperscript{12,13,14}, 
Franziska Mathis-Ullrich\textsuperscript{10},
Dwight Meglan\textsuperscript{15},
Vitor Mendes Pereira\textsuperscript{16},
Mouloud Ourak\textsuperscript{17},\\
Matteo Pantano\textsuperscript{1,6}, 
S.M.H. Sadati\textsuperscript{1,18},
Alice Taylor-Gee\textsuperscript{1},
Tom Vercauteren\textsuperscript{1},
Phil White\textsuperscript{19,20,21}, \\
Alejandro Granados\textsuperscript{1},
Thomas C Booth\textsuperscript{1,13,21}\thanks{Corresponding author: thomas.booth@kcl.ac.uk}
\\[1em]
\textsuperscript{1}School of Biomedical Engineering \& Imaging Sciences, Kings College London, London, UK\\
\textsuperscript{2}King’s Technology Evaluation Centre (KiTEC), Kings College London, London, UK\\
\textsuperscript{3}Patient \& Public Involvement, Kings College London, London, UK\\
\textsuperscript{4}Department of Interventional Neuroradiology, Rothschild Foundation Hospital, Paris, France\\
\textsuperscript{5}Department of Interventional Neuroradiology, North Bristol NHS Trust, Bristol, UK\\
\textsuperscript{6}Siemens Healthineers AG, Erlangen, Germany\\
\textsuperscript{7}Pattern Recognition Lab, FAU Erlangen-Nürnberg, Erlangen, Germany\\
\textsuperscript{8}Department of Neurology, University of Texas Rio Grande Valley, Edinburg, TX, USA\\
\textsuperscript{9}Society of Vascular and Interventional Neurology, Minneapolis, MN, USA\\
\textsuperscript{10}Department of Artificial Intelligence in Biomedical Engineering, FAU Erlangen-Nürnberg, Erlangen, Germany\\
\textsuperscript{11}Stroke Association, London, UK\\
\textsuperscript{12}European Society of Minimally Invasive Neurological Therapy, Zurich, Switzerland\\
\textsuperscript{13}Department of Neuroradiology, Kings College Hospital, London, UK\\
\textsuperscript{14}Department of Neuroradiology, Queens Hospital Romford, London, UK\\
\textsuperscript{15}Exemplar Devices LLC, Beavercreek, OH, USA\\
\textsuperscript{16}St. Michael's Hospital, Toronto, Canada\\
\textsuperscript{17}KU Leuven University, Leuven, Belgium\\
\textsuperscript{18}Queen Mary University of London, London, UK\\
\textsuperscript{19}Translational \& Clinical Research Institute, Newcastle University, Newcastle, UK\\
\textsuperscript{20}Newcastle upon Tyne Hospitals NHS Foundation Trust, Newcastle, UK\\
\textsuperscript{21}United Kingdom Neurointerventional Group, London, UK
}

\begin{document}
\maketitle

\paragraph{START contributors.}
Frédéric Clarençon (Department of Interventional Neuroradiology, Sorbonne Université, APHP, Pitié-Salpêtrière Hospital, Paris, France);
Adnan Siddiqui (Department of Neurosurgery, State University of New York at Buffalo, Buffalo, NY, USA);
David Bell (Remedy Robotics, San Francisco, CA, USA);
Nikola Fischer, Kawal Rhodes, Christos Bergeles (School of Biomedical Engineering \& Imaging Sciences, Kings College London, UK).

\paragraph{Society endorsement.}
Society of Vascular and Interventional Neurology (USA);
United Kingdom Neurointerventional Group.

\begin{abstract}

    While we are making progress in overcoming infectious diseases and cancer, one of the major medical challenges of the mid-21st century will be the rising prevalence of stroke. Occlusions in large vessels are especially debilitating, yet effective treatment—needed within hours to achieve best outcomes—remains limited due to geographic accessibility. One solution for improving timely access to mechanical thrombectomy in geographically diverse populations is the widespread deployment of robotic surgical systems. Artificial intelligence (AI) assistance may enable the safe and effective upskilling of operators in this emerging therapeutic delivery approach.
    
    Our aim was to establish consensus frameworks for developing and validating AI-assisted robots for thrombectomy. Objectives included standardizing effectiveness metrics and defining reference testbeds across in silico, in vitro, ex vivo, and in vivo environments. To achieve this, we convened experts in neurointervention, robotics, data science, health economics, policy, statistics, and patient advocacy. Consensus was built through an incubator day, a Delphi process, and a final Position Statement.
    
    We identified that the four essential testbed environments each had distinct validation roles. Realism requirements vary: simpler testbeds should include realistic vessel anatomy compatible with guidewire and catheter use, while standard testbeds should incorporate deformable vessels. More advanced testbeds should include blood flow, pulsatility, and disease features such as atheromatous plaques.
    
    There are two macro-classes of effectiveness metrics: one for in silico, in vitro, and ex vivo stages focusing on technical navigation (e.g., path-following error), and another for in vivo stages, focused on clinical outcomes (e.g., modified treatment in cerebral infarction scores). Patient safety is central—and not a barrier—to this technology’s development. One requisite patient safety task needed now is to correlate in vitro measurements to in vivo complications.
\end{abstract}


\section{Introduction}\label{sec1}

    The annual number of strokes and stroke-related deaths increased by 70\% and 43\%, respectively, from 1990 to 2019, making stroke the second-leading cause of death and the third highest contributor to the burden of disease worldwide~\cite{Feigin2021, Vos2020}. Mechanical thrombectomy (MT) has emerged as a standard treatment for acute ischemic stroke resulting from large vessel occlusion, providing improved functional outcomes when compared with medical treatment alone~\cite{Nogueira2018, Goyal2016, Bendszus2023, Albers2018}.

    In cases of ischemic stroke, the timing of intervention from symptom onset plays a critical role in the effectiveness of MT. Notably, the efficacy of MT diminishes significantly after 7.3\,\si{\hour} from stroke onset for non-stratified patients~\cite{Saver2016}. Despite recent evidence suggesting a growing proportion of stroke patients qualifying for MT~\cite{Bendszus2023}, and despite the potential benefits of MT, there is a significant gap in treatment accessibility seen within all countries~\cite{SSNAP2023, McMeekin2017, Asif2023}. Even in high income countries a significant gap exists. In the UK, for example, only 3.1\% of stroke admissions received MT in 2023, despite 10\% being eligible~\cite{SSNAP2023, McMeekin2017}. Patients arriving from remote stroke centers are less likely to receive MT if the door-to-door time is $<$3\,\si{\hour}, which occurs in 45\% of admissions~\cite{Zhang2021}. Similar findings have been seen in the US where the probability of undergoing MT decreases by 1\% for each additional minute of transfer time over an ideal transfer time of 1\,\si{\hour}~\cite{Regenhardt2018}.

    Challenges associated with MT include occasional complications such as vessel perforations (1\%), procedure-related vessel dissections (2\%), and distal embolization of thrombus (9\%)~\cite{Berkhemer2015}. Beyond patient safety, operators and their teams face the risk of potentially high cumulative doses of x-ray radiation from angiography, which pose risks of cancer and cataracts~\cite{Klein2009}. Although current radiation protection practices help minimize exposure, some measures, such as wearing heavy protective equipment, can lead to orthopedic complications~\cite{Ho2007, Madder2017}. One proposed solution that mitigates these challenges whilst ensuring timely access to MT for diverse populations, involves the deployment of cost effective robotic surgical systems, which could be strategically positioned in hospitals nationwide and then operated remotely from a central hub by experts, operated with artificial intelligence (AI) assistance by competent but non-expert operators, or even autonomously~\cite{Crinnion2022}. This would greatly increase access to treatment, where currently 31.2\% of the US population have no access to an interventional radiologist within their county~\cite{Ahmad2024}. Robotic interventions executed by trained operators offer several advantages, including the potential elimination of operator tremors, reduced radiation exposure, and enhanced procedural precision, thereby minimizing complications~\cite{Riga2010}. Robotically actuated systems have demonstrated successful outcomes in neuroendovascular interventions~\cite{Pereira2020, Cancelliere2022}. However, robotically actuated systems which employ a controller-operator structure may lead to high cognitive workloads with the potential for human error~\cite{vanDijk2023}, although this has not yet been proven for endovascular robotics.

    Integrating AI techniques with robotic actuation systems has emerged as a promising approach to reduce risk and increase clinical acceptance of robotic surgical systems. Machine learning (ML) has seen rapid advancement offering potential solutions for assisted navigation of guidewires and catheters during MT~\cite{Sarker2021}. ML algorithms may provide assistance or autonomy in navigation, mitigating the challenges associated with full manual control including human factors such as fatigue and loss of focus, potentially enhancing procedural safety and efficiency~\cite{Mirnezami2018}. Additionally, autonomous or assisted navigation might upskill local generalist operators rather than relying on tele-operation by experts from central hubs. Therefore, AI assistance democratizes the ability of all operators, while minimizing clinical differences in operator technique and potentially judgement too.

    While several studies have explored the integration of ML in automating catheter and guidewire manipulation for endovascular interventions, a recent systematic review highlighted the lack of high-level evidence to demonstrate that AI-based autonomous navigation of catheters and guidewires in any endovascular intervention is non-inferior or superior to manual procedures~\cite{Robertshaw2023, Howick2011}. The review found that the field has not surpassed an experimental proof-of-concept stage with a technology readiness level (TRL) of 3~\cite{Mankins1995}, which corresponds to active research and development efforts that involve both analytical studies to place the technology in context, and laboratory-based experiments to validate the predictions made during the conceptual phase (TRL 2)~\cite{Mankins1995}. Furthermore, the review found that there are no standardized \textit{in silico}, \textit{in vitro}, \textit{ex vivo} or \textit{in vivo} experimental reference standard testbeds, nor are there standardized effectiveness metrics, meaning that comparison of studies quantitatively is of limited value.

    To increase the TRL of the field, and move towards demonstrating any benefits of ML for the autonomous navigation of MT equipment, it is vital to establish reference and reporting standards for robotic MT. Such frameworks will also be suitable for a wide range of MT development regardless of whether robots or ML are employed.
    
    The purpose of this Position Statement was to establish, through consensus, reference and reporting standard frameworks to be used for the development and validation of robotic MT, both with and without AI. Objectives were to standardize effectiveness metrics (including determining standardized units of measurement) and reference standard testbeds for \textit{in silico}, \textit{in vitro}, \textit{ex vivo} and \textit{in vivo} environments.

\section{Methods}\label{sec2}

    \subsection{Incubator Meeting}

        A multidisciplinary group comprising healthcare professionals, academia and industry experts, and patient advocacy representatives, convened for an incubator day on AI-informed robotics for MT in ischemic stroke. The first meeting took place in London, UK, in April 2024, with the option to attend virtually to allow for international attendance. There were twenty-one expert attendees; invited based on their expertise in neuroendovascular procedures, robotics, data science, health economics, healthcare policy, statistics and patient advocacy.
        
        Current endovascular testbed reference standards and effectiveness metrics, for \textit{in silico}, \textit{in vitro}, \textit{ex vivo} and \textit{in vivo} environments were reviewed and discussed among the group, where it was concluded that standardization among these areas is needed to improve AI research outcomes~\cite{Robertshaw2023}. Additionally, current MT clinical effectiveness metrics were reviewed~\cite{Goyal2016,Saver2016}, and consideration was given to what metrics could be used in early TRL research to give the highest likelihood of improved clinical outcomes at later translational stages~\cite{Crossley2019}.

        Based on this meeting, a consensus was reached that a Delphi study should be conducted in the first instance to establish standards for neuroendovascular testbeds and effectiveness metrics across \textit{in silico}, \textit{in vitro}, \textit{ex vivo} and \textit{in vivo} environments for autonomous robotics in MT. Following this, it was agreed that the consensus report of a Delphi study~\cite{Prinsen2014}, would form the basis of the Position Statement~\cite{Europarat2002}. The Position Statement is particularly well-suited to promoting discussion on emerging topics and identifying evidence gaps, as well as highlighting current strengths and limitations in the field. Together this would provide the foundation for producing a future multi-stakeholder guideline~\cite{Europarat2002} that builds upon the current Position Statement.

    \subsection{Delphi Method}

        A Delphi consensus is commonly used to gain agreement across a panel of experts, and was used here to provide anonymity of the panel, while ensuring that each participant had an equal possibility to provide and change their opinion during the course of the process~\cite{Keeney2001, Hasson2000}. All members of the group that attended the incubator meeting were invited to be Delphi panelists.

\begin{figure}[]
            \centering
            \includegraphics[width=9cm]{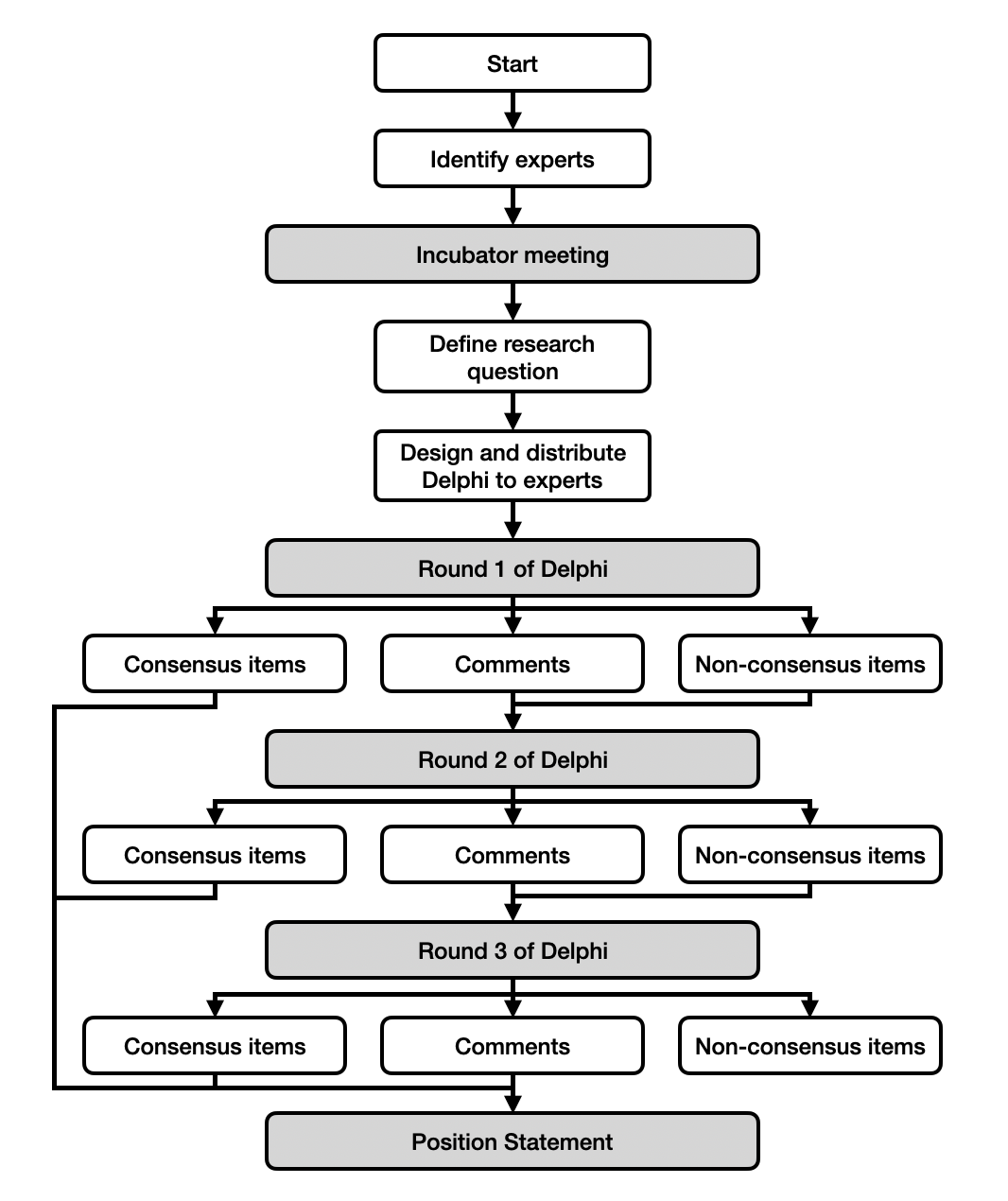}
            \caption{ Flowchart for methods followed in this study, including incubator meeting, three rounds of Delphi, and final Position Statement.}
            \label{fig:delphi}
        \end{figure}

        The Delphi exercise consisted of three rounds~\cite{Sinha2011}. The process followed is shown in Figure~\ref{fig:delphi}. The first round was sent by email to all panelists in May 2024. For all rounds, a reminder was sent two weeks later to individuals who did not respond to the initial invite. The second round was distributed in July 2024. Any questions that reached 80\% consensus in the previous round were considered as finalized and were removed from the questionnaire. The remaining questions which had less than 80\% consensus were included in the questionnaire and contributors were informed of the current percentage of agreement from the previous round. New options were also added to the questionnaire on the basis of panelist feedback through free-text comments from the previous round. The same process was followed for round three, which was sent out in September 2024.

        The survey was split into four sections. In section one, panelists were asked to provide baseline information about their experience with MT. In section two, participants evaluated various aspects of robotic MT, including benefits, risks, and barriers to standardization across different phases of AI-integrated MT. Questions covered various configurations of robotic MT, including systems both with and without AI. Panelists were asked to consider each factor independently, without assuming other factors were present, and were encouraged to propose additional benefits, risks, or factors requiring standardization.

\begin{figure}[]
            \centering
            \includegraphics[width=14cm]{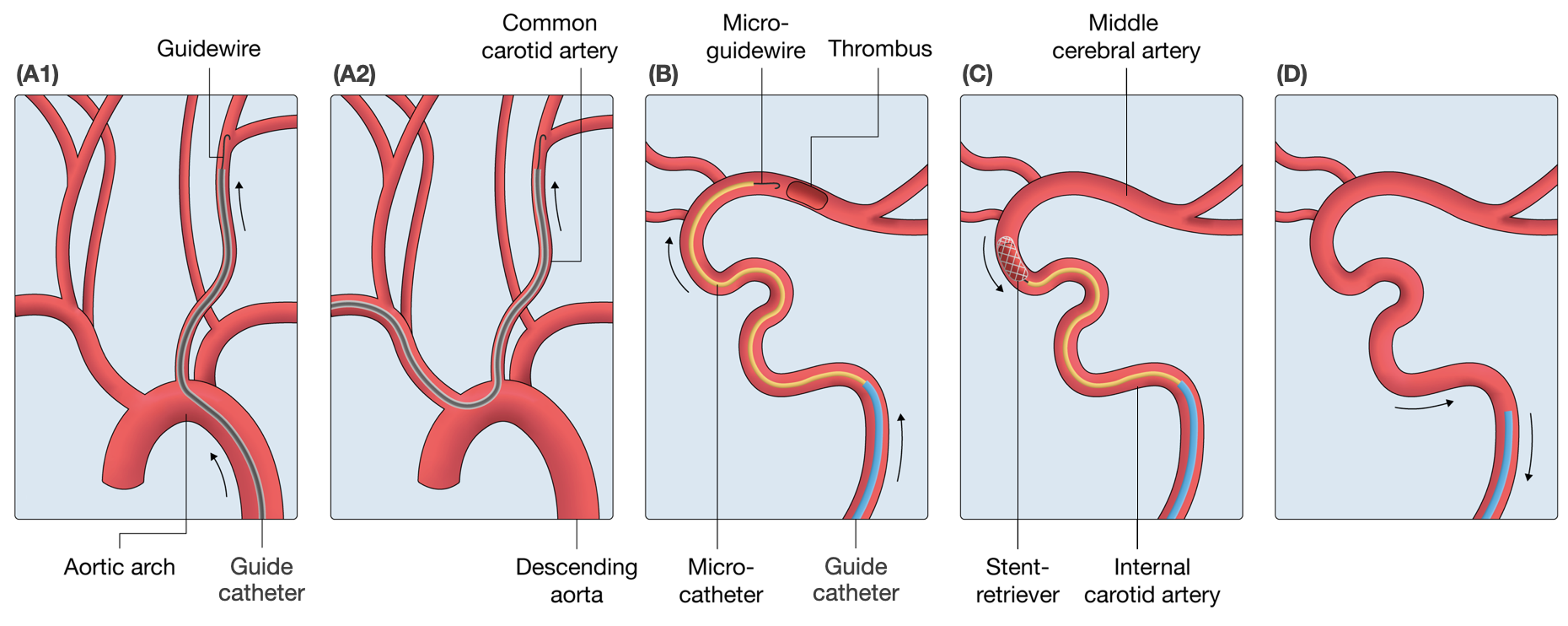}
            \caption{ Defined phases of MT intervention: (A1) primary access (femoral artery), (A2) primary access (radial artery), (B) secondary access, (C) treatment (stent shown here, but could also be an aspiration catheter) (D) removal of navigation equipment (and access closure). The ``navigation phases" are considered to be (A) primary and (B) secondary access.}
            \label{fig:phases}
        \end{figure}

        Section three focused on neuroendovascular testbed reference standards (also referred to as benchmarks). Panelists separately evaluated different developmental stages of the innovation lifecycle which consisted of \textit{in silico}, \textit{in vitro}, \textit{ex vivo} (focusing on human cadavers), and \textit{in vivo} (testbeds, by definition, must be non-human) environments. Within each developmental stage, defined phases~\cite{Robertshaw2025} of anterior circulation MT (Figure~\ref{fig:phases}) were considered:

        \begin{itemize}
            \item A1) Primary access, using guidewire and guide catheter: femoral artery to common (or internal) carotid artery
            \item A2) Primary access, using guidewire and guide catheter: radial artery to common (or internal) carotid artery
            \item B) Secondary access, refers to a phase following primary access, micro-catheter and micro- guidewire/ or aspiration catheter and wire: internal carotid artery to cerebral artery
            \item C) Treatment, using stent retriever or aspiration catheter
            \item D) Removal of navigation equipment, and access closure
        \end{itemize}

        The panelists rated the importance of specific testbed features within each environment type, such as realistic vessel anatomy, deformable vessels, simulated blood flow, pulsatility, respiration, use of both catheter and guidewire, and representation of the diseased vessel (e.g., atherosclerotic plaques).
        
        Section four of the survey assessed the utility of various effectiveness metrics in each type of environment to evaluate robotic MT procedures. The initially-proposed MT navigation effectiveness metrics for \textit{in silico}, \textit{in vitro}, \textit{ex vivo} and \textit{in vivo} environments were derived from neuroendovascular simulation research~\cite{Crossley2019} as well as a recent systemic review that captured all metrics published so far in the field of endovascular AI assistance~\cite{Robertshaw2023}. Panelists rated the following metrics for their utility as standards to be considered predominantly for the two ``navigation phases" ((A) primary and (B) secondary access):

        \begin{itemize}
            \item Success rate: how many times can the robot reach the target in a given number of evaluations
            \item Number of procedural phases: number of MT procedural phases completed in a navigation attempt (Figure~\ref{fig:phases}), as defined by~\cite{Crossley2019}
            \item Number of failures: number of failures made during a navigation attempt (e.g., wrong branch catheterization)
            \item Number of handling errors: number of MT handling errors made in a navigation attempt, as defined by~\cite{Crossley2019}
            \item Procedure time
            \item Path length: length of path taken by the device tip from insertion point to target
            \item Path following error: difference between the catheter tip path to the vessel centerline
            \item Instrument tip speed
            \item Instrument tip acceleration
            \item Instrument tip contact forces: mean and maximum contact forces applied to the tip of the instrument during a navigation attempt
            \item Instrument base contact forces: mean and maximum contact forces applied to the base of the instrument during a navigation attempt
            \item Vessel walls contact forces: mean and maximum contact forces applied to the vessel walls during a navigation attempt
            \item Fluoroscopy time
            \item Contrast agent volume
            \item Number of guidewire tip touches on vessel walls
        \end{itemize}

        Whilst we considered what an appropriate non-human \textit{in vivo} testbed might consist of during the Delphi process, we also noted the paucity of non-human \textit{in vivo} testbed and effectiveness metrics described in the literature for MT development, and the well-defined clinical assessment metrics published in numerous MT trials. These clinical assessment metrics covered criteria such as modified treatment in cerebral infarction (mTICI) scores~\cite{Zaidat2013}, extended treatment in cerebral infarction (eTICI) scores~\cite{Liebeskind2019}, first-pass success rates, vessel perforation, dissection, hemorrhage, distal embolization, and procedural failure~\cite{Berkhemer2015}. These MT clinical effectiveness metrics are the most important metrics for translation as they have been used to derive the level 1 evidence~\cite{Howick2011} for MT~\cite{Goyal2016,Saver2016}. The metrics relate to safety and efficacy following an entire MT procedure~\cite{Goyal2016,Saver2016} and not navigation steps. Therefore, in our Delphi process, \textit{in vivo} effectiveness metric questions focused on clinical effectiveness and not navigation. In contrast, navigation research may be more suitable for \textit{in silico}, \textit{in vitro}, and \textit{ex vivo} environments and here effectiveness metric questions focused on navigation during the Delphi process.

        Answers to questions from all four sections were either binary (yes or no) or used a 5-point Likert scale (e.g., strongly agree, agree, neither agree nor disagree, disagree, strongly disagree). All questions had an additional `unsure' option and space for free-text comments for panelists to make suggestions for future round or to provide comments. Those who selected `unsure' for an answer (reflecting lack of expertise in that particular domain) were not included in the consensus calculation for that particular question.

        Comparative statistics included the chi-square test. Significance was set at $p = 0.05$. Statistical analyses were conducted using R version 4.3 (R Foundation for Statistical Computing, Vienna, Austria).

\section{Results}\label{sec3}

    \subsection{Delphi Panel}

        Out of the 21 experts who attended the incubator meeting, 20 opted to participate in the Delphi (one charity representative attendee considered the specialist Delphi exercise beyond their technical capability; but the charity has contributed to the final Position Statement). Two more panelists were added to the Delphi consensus based on recommendations following the incubator meeting. Out of the 22 experts that took part in the Delphi study, 22/22 (100\%) completed Rounds 1, 2 and 3. The majority of panelists were academic researchers (64\%, 14/22), followed by clinicians (32\%, 7/22), data scientists/engineers (23\%, 5/22), industry representatives (14\%, 3/22), patient representatives (9\%, 2/22) and statisticians (5\%, 1/22) (panelists were able to identify as being an expert from more than one background). European panelists made up 82\% (18/22) of participants, while 18\% (4/22) were from North America. The majority of panelists had observed or performed MT procedures (59\%, 13/22).

    \subsection{Robotic MT and AI}

        Almost all respondents recognized the potential benefits of robotic MT, with 95\% (21/22) endorsing robotic MT without AI, and 100\% (22/22) endorsing robotic MT with AI assistance in the broadest sense. Notably, AI-assisted robotic MT for navigation was endorsed as a beneficial tool for two broad use cases: for use by neurointerventional experts in a tele-operated capacity (91\%, 20/22), and for use by general interventionalists (i.e. those performing endovascular procedures but not endovascular procedures related to ischemic or hemorrhagic stroke) who would be upskilled by the technology (95\%, 21/22). Fully autonomous robotic MT elicited support of 86\% (19/22) by the end of the Delphi exercise. Participants unanimously agreed that it was important to standardize neuroendovascular testbeds and effectiveness metrics for \textit{in silico}, \textit{in vitro}, \textit{ex vivo}, and \textit{in vivo} environments. 

\begin{figure}[htb!]
            \centering
            \includegraphics[width=14cm]{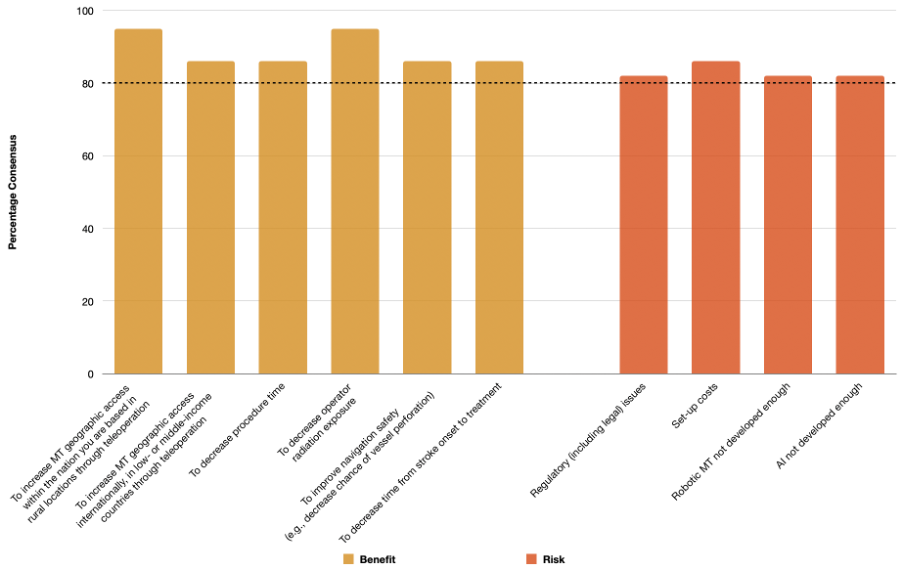}
            \caption{ Consensually agreed benefits and risks of robotic mechanical thrombectomy (MT), both with and without artificial intelligence (across all developmental stages of robotic MT from \textit{in silico} to \textit{in vivo}). Percentage agreement for each benefit and risk was taken from the round where consensus was first reached. Benefits and risks of robotic MT that did not reach consensus in any round can be found in Supplementary Table~\ref{tab:non_consensus_bens_risks}.}
            \label{fig:benefits_risks}
        \end{figure}

        Figure~\ref{fig:benefits_risks} shows the consensually agreed potential benefits and risks of robotic MT, both with and without AI (across all types of robotic MT); Supplementary Table~\ref{tab:non_consensus_bens_risks} shows the benefits and risks that did not reach consensus. One notable benefit endorsed during the Delphi exercise was the ability of robotic MT to increase access in rural locations via tele-operation, which was supported by 95\% (21/22) of participants. Additional perceived benefits included reducing operator radiation exposure (95\%, 21/22), decreasing procedure time (86\%, 19/22), decreasing time from stroke onset to treatment (86\%, 19/22), enhancing navigation safety (86\%, 19/22), and international access for low- and middle-income countries (86\%, 19/22). However, a reduction in patient radiation and contrast exposure was not perceived to be a potential benefit from employing robotic MT (with or without AI). Legal and regulatory concerns (82\%, 18/22), insufficient development of robotic MT (82\%, 18/22), set-up costs (86\%, 19/22), and insufficient development of AI (82\%, 18/22), were considered to be risks associated with AI-based MT navigation not succeeding. Ethical concerns, patient safety, and societal concerns were not considered to be risks \textit{per se} to AI-based MT navigation succeeding (Note: in contrast, patient safety was considered the most important consideration in the Position Statement overall and is discussed below. Similarly, ethical concerns related to implementation warranted a future Position Statement).

    \subsection{Neuroendovascular Testbeds}

\begin{figure}[htb!]
            \centering
            \includegraphics[width=14cm]{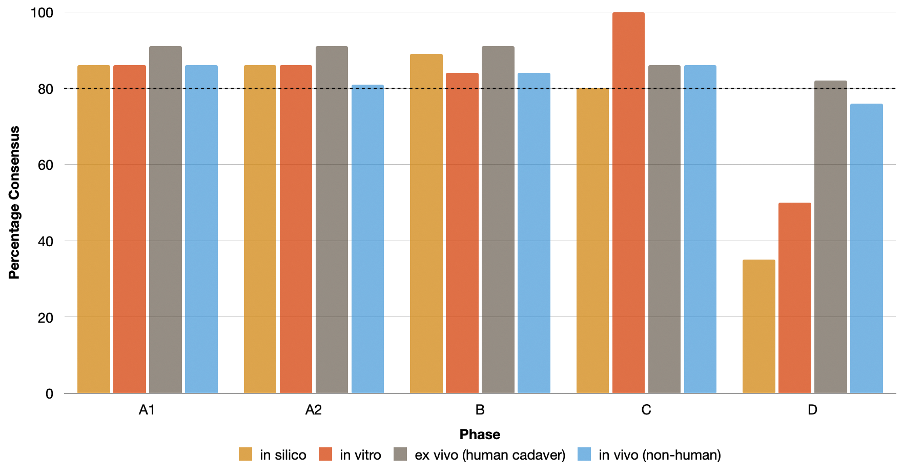}
            \caption{ Consensus of which mechanical thrombectomy (MT) phases are effective for the development of robotic MT both with and without artificial intelligence (across all developmental stages of robotic MT from \textit{in silico} to \textit{in vivo}), for testbeds from each developmental stage. Percentage agreement for each phase was taken from the round where consensus was first reached.}
            \label{fig:phases_results}
        \end{figure}

        The panel considered whether testbeds at each developmental stage (\textit{in silico} through to \textit{in vivo}) were appropriate environments to benchmark robotic MT, both with and without AI (Fig.~\ref{fig:phases_results}). The MT phases (A1, A2, B, C, D) were considered at each developmental stage. Effective design of robotic MT during phases A1, A2, B and C was considered achievable using all four developmental stage testbeds. It was noted that there was no relationship between these phases and developmental stage (A1 ($\chi^2 (1,N=86)=0.37, p=0.95$), A2 ($\chi^2 (1,N=87)=0.90, p=0.83$), B ($\chi^2 (1,N=79)=0.67, p=0.88$) or C ($\chi^2 (1,N=86)=2.30, p=0.51$)). In contrast, developing robotic MT to achieve phase D (removal of instruments and access closure) was only considered effective in \textit{ex vivo} (cadaver) environments (82\%, 18/22).
        
        Table~\ref{tab:consensus_factors} lists the experimental factors considered important for testbeds from each developmental stage (both with and without AI). Experimental factors that did not reach consensus in any round can be found in Supplementary Table~\ref{tab:non_consensus_factors}. Realistic anatomy (including diseased vessels), vessel deformability and combined use of catheters and guidewires together, were considered important factors to be incorporated into all four developmental stage testbeds. Simulated blood flow was also supported in testbeds, while simulated respiration was not (note that blood flow, pulsatility and respiration would not need to be simulated \textit{in vivo}).

        \begin{table}[h!]
            \footnotesize
            \centering
            \caption{Consensus of which experimental factors were considered important for each developmental stage testbed (both with and without artificial intelligence). Percentage agreement for each factor was taken from the round where consensus was first reached. Note that blood flow and pulsatility would not need to be simulated \textit{in vivo}. Experimental factors that did not reach consensus in any round can be found in Supplementary Table~\ref{tab:non_consensus_factors}.}
            \label{tab:consensus_factors}
            \begin{tabular}{p{5.2cm}cccc}
                \hline
                \textbf{Experimental Factor} & \textbf{\textit{in silico}} & \textbf{\textit{in vitro}} & \textbf{\textit{ex vivo}} & \textbf{\textit{in vivo} (non-human)} \\
                \hline
                Realistic vessel anatomy / Human-like vessel anatomy & 84\% ($16/19$) & 95\% ($18/19$) & 84\% ($16/19$) & 95\% ($21/22$) \\
                Deformable vessels & 89\% ($17/19$) & 84\% ($16/19$) & 84\% ($16/19$) & 84\% ($16/19$) \\
                Simulated blood flow & 82\% ($18/22$) & 84\% ($16/19$) & 86\% ($18/21$) & --- \\
                Simulated pulsatility & 86\% ($18/21$) & 95\% ($19/20$) & --- & --- \\
                Use of both catheter and guidewire for navigation as opposed to single item & 95\% ($18/19$) & 95\% ($18/19$) & 89\% ($17/19$) & 89\% ($17/19$) \\
                Diseased vessel containing atherosclerotic plaques & 90\% ($19/21$) & 86\% ($19/22$) & 80\% ($16/20$) & 90\% ($18/20$) \\
                \hline
            \end{tabular}
        \end{table}
    
    \subsection{Effectiveness Metrics}

        \begin{table}[h!]
            \footnotesize
            \centering
            \caption{Consensus on important effectiveness measures for each developmental stage (with and without AI), as well as during \textit{in vivo} clinical-like assessment. Percentage agreement for each factor was taken from the round where consensus was first reached. Measures that did not reach consensus in any round can be found in Supplementary Table~\ref{tab:non_consensus_performance_measures}.}
            \label{tab:consensus_performance_measures}
            \begin{tabular}{p{4.7cm} p{2cm} p{2cm} p{2cm} p{2cm}}
                \hline
                \textbf{Effectiveness Measure} & \textbf{\textit{in silico}} & \textbf{\textit{in vitro}} & \textbf{\textit{ex vivo}} & \textbf{\textit{in vivo}} \\
                \hline
                Success rate & 100\% ($20/20$) & 100\% ($19/19$) & 100\% ($19/19$) & --- \\
                Number of phases or steps \cite{Crossley2019} & 84\% ($16/19$) & 90\% ($19/21$) & 84\% ($16/19$) & --- \\
                Number of failures (e.g., wrong branch catheterization) & 95\% ($19/20$) & 89\% ($17/19$) & 89\% ($17/19$) & --- \\
                Number of handling errors made \cite{Crossley2019} & 84\% ($16/19$) & 89\% ($17/19$) & 95\% ($18/19$) & --- \\
                Procedure time & 85\% ($17/20$) & 89\% ($17/19$) & 86\% ($19/22$) & --- \\
                Path following error & 81\% ($17/21$) & 81\% ($17/21$) & --- & --- \\
                Path length & --- & 81\% ($17/21$) & --- & --- \\
                Instrument tip speed & --- & 80\% ($16/20$) & --- & --- \\
                Contact forces at instrument tip & 89\% ($16/18$) & 82\% ($14/17$) & 86\% ($18/21$) & --- \\
                Contact forces at instrument base & 83\% ($15/18$) & 82\% ($18/22$) & 90\% ($18/20$) & --- \\
                Contact forces on vessel walls & 84\% ($16/19$) & 86\% ($19/22$) & 86\% ($18/21$) & --- \\
                Fluoroscopy time & 90\% ($19/21$) & --- & 80\% ($16/20$) & --- \\ \hline
                mTICI raw score & --- & --- & --- & 92\% ($11/12$) \\
                eTICI raw score & --- & --- & --- & 83\% ($10/12$) \\
                “Successful recanalization” first pass (mTICI $\geq$ 2b, after 1 pass) & --- & --- & --- & 91\% ($10/11$) \\
                “Complete recanalization” first pass (mTICI $\geq$ 2c, after 1 pass) & --- & --- & --- & 91\% ($10/11$) \\
                “Successful recanalization” (mTICI $\geq$ 2b, after $\geq$1 pass) & --- & --- & --- & 82\% ($9/11$) \\
                “Complete recanalization” (mTICI $\geq$ 2c, after $\geq$1 pass) & --- & --- & --- & 91\% ($10/11$) \\
                Vessel perforation & --- & --- & --- & 100\% ($18/18$) \\
                Vessel dissection & --- & --- & --- & 94\% ($16/17$) \\
                Intracranial hemorrhage & --- & --- & --- & 94\% ($16/17$) \\
                Distal embolization & --- & --- & --- & 100\% ($16/16$) \\
                Procedural failure & --- & --- & --- & 94\% ($17/18$) \\
                \hline
            \end{tabular}
        \end{table}

        The effectiveness measures selected by consensus for each developmental stage are shown in Table~\ref{tab:consensus_performance_measures}. Measures that did not reach consensus in any round can be found in Supplementary Table~\ref{tab:non_consensus_performance_measures}.
        
        Four conclusions can be drawn from the Delphi exercise. First, we propose that there are multiple effectiveness metrics which should be measured during \textit{in silico}, \textit{in vitro}, and \textit{ex vivo} developmental stages for navigation research. Navigation metrics are related to process (success rate, number of phases or steps, number of failures, number of handling errors made), duration (procedure time), and contact forces (at instrument tip, at instrument base, on vessel walls). Similarly, we propose that there are multiple clinical effectiveness metrics that should be measured during \textit{in vivo} research and are suited for measurement of the entire MT procedure; these can be applied to non-human as well as clinical (human) research.

        Second, of the effectiveness metrics considered to be essential for navigation research at all developmental stages, success rate, defined as the number of successful attempts in a given number of evaluations, was the most important metric reaching 100\% (20/20) at every stage.

        Third, four effectiveness metrics were not considered to be essential at all developmental stages (path following error, path length, instrument tip speed and fluoroscopy time) by the Delphi panel. However, they were considered important at certain developmental stages (e.g. fluoroscopy time was considered important during \textit{in silico} and \textit{ex vivo} testing, but not \textit{in vitro} testing). Additionally, several effectiveness metrics were not considered to be essential across any developmental stage. These included volume of the contrast agent used, instrument tip acceleration, and number of contacts between the guidewire tip and vessel wall.

        Fourth, all \textit{in vivo} clinical assessment metrics assessed, including efficacy TICI scores, complications and procedural failure, reached consensus.

    \subsection{Consensus Summary}

        Based on the consensus agreements reached in the Delphi exercise, we summarized and organized the information by developmental stage for user reference (Table~\ref{tab:combined_consensus}).

        \begin{table}[htbp]
            \footnotesize
            \centering
            \caption{Combined consensus recommendations for testbeds and effectiveness measures across developmental stages. We have also given units of measure. MT: mechanical thrombectomy; eTICI: extended treatment in cerebral infarction; mTICI: modified treatment in cerebral infarction.}
            \label{tab:combined_consensus}
            \begin{tabular}{p{5cm}cccc}
                \hline
                \textbf{Item} & \textbf{\textit{in silico}} & \textbf{\textit{in vitro}} & \textbf{\textit{ex vivo}} & \textbf{\textit{in vivo}} \\
                \hline
                \textbf{MT Navigation Phases} & & & & (non-human) \\
                Phase A1 & \checkmark & \checkmark & \checkmark & \checkmark \\
                Phase A2 & \checkmark & \checkmark & \checkmark & \checkmark \\
                Phase B & \checkmark & \checkmark & \checkmark & \checkmark \\
                Phase C & \checkmark & \checkmark & \checkmark & \checkmark \\
                Phase D & --- & --- & \checkmark & --- \\
                \hline
                \textbf{Experimental Considerations} & & & & (non-human)\\
                Realistic vessel anatomy & \checkmark & \checkmark & \checkmark & \checkmark \\
                Deformable vessels & \checkmark & \checkmark & \checkmark & \checkmark \\
                Simulated blood flow & \checkmark & \checkmark & \checkmark & --- \\
                Simulated pulsatility & \checkmark & \checkmark & --- & --- \\
                Use of catheter \& guidewire & \checkmark & \checkmark & \checkmark & \checkmark \\
                Diseased vessel (plaques) & \checkmark & \checkmark & \checkmark & \checkmark \\
                \hline
                \multicolumn{5}{l}{\textbf{Effectiveness Measures}} \\
                Success rate (\%)                            & \checkmark & \checkmark & \checkmark & --- \\
                Number of phases or steps                    & \checkmark & \checkmark & \checkmark & --- \\
                Number of failures                           & \checkmark & \checkmark & \checkmark & --- \\
                Handling errors made                         & \checkmark & \checkmark & \checkmark & --- \\
                Procedure time (s)                           & \checkmark & \checkmark & \checkmark & --- \\
                Path following error (\%)                    & \checkmark & \checkmark & ---         & ---         \\
                Path length (m)                              & ---         & \checkmark & ---         & ---         \\
                Instrument tip speed (m/s)                   & ---         & \checkmark & ---         & ---         \\
                Contact forces at tip (N)                    & \checkmark & \checkmark & \checkmark & --- \\
                Contact forces at base (N)                   & \checkmark & \checkmark & \checkmark & --- \\
                Contact forces on vessel walls (N)           & \checkmark & \checkmark & \checkmark & --- \\
                Fluoroscopy time (s)                         & \checkmark & ---         & \checkmark & --- \\
                \hline
                \multicolumn{5}{l}{\textbf{Effectiveness measures (focus on clinical assessment of entire MT procedure; human and non-human)}} \\
                mTICI (modified treatment in cerebral infarction) raw score               & --- & --- & --- & \checkmark \\
                eTICI (extended treatment in cerebral infarction) raw score              & --- & --- & --- & \checkmark \\
                “Successful recanalization” first pass rate (mTICI 2b or higher, 1 pass) & --- & --- & --- & \checkmark \\
                “Complete recanalization” first pass rate (mTICI 2c or higher, 1 pass)   & --- & --- & --- & \checkmark \\
                “Successful recanalization” rate (mTICI 2b+, $\geq$1 passes)                  & --- & --- & --- & \checkmark \\
                “Complete recanalization” rate (mTICI 2c+, $\geq$1 passes)                    & --- & --- & --- & \checkmark \\
                Vessel perforation                                                        & --- & --- & --- & \checkmark \\
                Vessel dissection                                                         & --- & --- & --- & \checkmark \\
                Intracranial hemorrhage                                                                & --- & --- & --- & \checkmark \\
                Distal embolization                                                       & --- & --- & --- & \checkmark \\
                Procedural failure                                                        & --- & --- & --- & \checkmark \\
            \end{tabular}
        \end{table}

\section{Discussion}\label{sec4}

    The Delphi consensus recommendations were synthesized by the panel and informed the following Positions. We first considered endovascular testbeds and effectiveness metrics in depth in keeping with the Position Statement objectives. In addition, panelists summarized additional perspectives related to AI-assisted robotic MT, to highlight key points and flag them for future analyses by START initiatives.

    \subsection{Endovascular Testbeds}

\begin{figure}[]
            \centering
            \includegraphics[width=12cm]{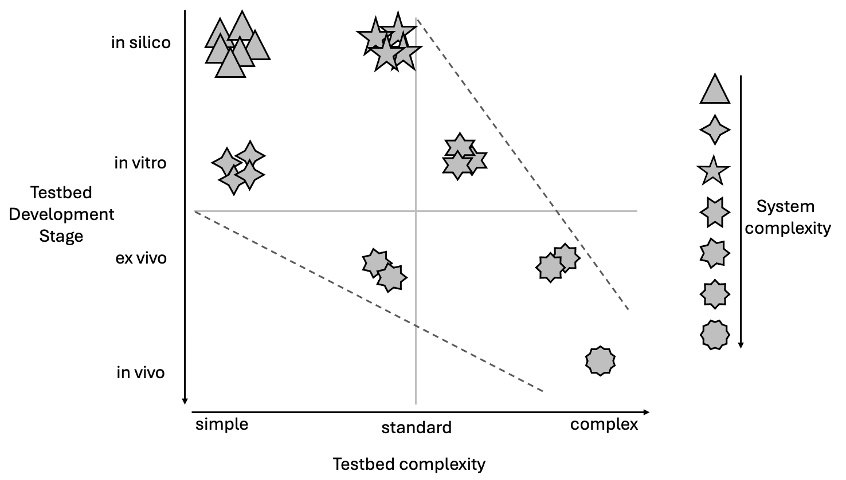}
            \caption{ Evolution of testbed complexity with feature complexity. Examples of system complexity—simple \textit{in silico}, simple \textit{in vitro}, standard \textit{in silico}, standard \textit{in vitro}, standard \textit{ex vivo}, complex \textit{ex vivo}, and complex \textit{in vivo} - are shown in Supplementary Figure~\ref{fig:simple_in_silico} through \ref{fig:complex_in_vivo}, respectively.}
            \label{fig:testbed1}
        \end{figure}

        The development process of robotic MT relies on four distinct testbed methodologies: \textit{in silico}, \textit{in vitro}, \textit{ex vivo}, and \textit{in vivo}, each serving specific evaluation and validation purposes. While \textit{in silico} and \textit{in vitro} testbeds serve as primary platforms for early-stage research, we acknowledge that implementing experimental factors such as simulated blood flow and pulsatility might be ideal as they are recommendations from the Delphi consensus. However, they present considerable challenges such as increased financial and logistical cost to develop computational or physical experimental testbeds. Introducing such complex elements at the outset could potentially impede innovation in this emerging field~\cite{Bolaos2021} – whilst we therefore recommend permissiveness at the outset with a `minimum viable product,' if the engineering capability and budget are not constrained, these physiological factors can be introduced at the outset. In this Position Statement we address these challenges by proposing a progressive approach that emphasizes accessibility for researchers during initial experimental stages while systematically incorporating additional features recommended from the Delphi consensus to enhance realism and allow more comprehensive experimental analyses. Early development of robotic MT systems, whether AI-assisted or not, focuses on evaluating specific features using simplified testbeds. As development progresses toward regulatory certification for clinical use, the complexity must gradually increases to match real interventional scenarios. Throughout the development cycle of an MT robotic system, testbeds transition from \textit{in silico} to \textit{in vivo}, with mandated complexity progressing from simple to complex configurations, as illustrated in Figure~\ref{fig:testbed1}. It is important to recognize that each development stage has inherent complexity limitations. For example, experiments conducted \textit{in silico} are unlikely to achieve full equivalence with those conducted \textit{in vivo}, while \textit{in vivo} testing cannot be simplified to control variables below certain inherent complexity levels.
        
        This approach creates an innovation funnel, which facilitates the rapid evaluation of multiple concepts at the beginning while ensuring the rigorous testing of the MT robotic systems at later stages. We define three different levels of complexity across the modality spectrum of \textit{in silico}, \textit{in vitro}, \textit{ex vivo}, and \textit{in vivo}: simple, standard, and complex:

        \begin{itemize}
            \item The simple level incorporates realistic vessel anatomies compatible with guidewire and catheter usage.
            \item The standard level builds upon this foundation and introduces deformable vessels. 
            \item The complex level achieves maximum realism by incorporating blood flow, pulsatility, and disease conditions (such as atheromatous plaques).
        \end{itemize}

        A typical evaluation process begins with accessible methods, including simple \textit{in silico} and \textit{in vitro} testbeds, and advances to more complex \textit{ex vivo}, and ultimately \textit{in vivo} experiments. To promote research advancement, maintaining a low entry barrier for comparable and reproducible evaluation is crucial. Open-source simulation frameworks and benchmarks, such as stEVE (simulated EndoVascular Environment,~\cite{Karstensen2024}), provide simple \textit{in silico} testbeds that enable reproducible evaluation and globally comparable results.

        These approaches can be further validated using simple \textit{in vitro} testbeds, featuring rigid transparent vessel phantoms with overhead camera imaging feedback, eliminating the absolute requirement for fluoroscopy systems. Commercial simulators like the Cathis (CATHI GmbH, Germany) or the VIST System (Mentice AB, Sweden) enable \textit{in silico} testing at standard and early complex levels. AI could also be used to generate anatomically complex \textit{in silico} testbeds, such as synthetic vascular models derived from 3D spline-based methods for training and validation of autonomous systems~\cite{Nader2024}. Similarly, \textit{ex vivo} testing can achieve standard and complex levels by incorporating flexible vessel systems and pulsatile pumps. Many essential components are commercially available, including sophisticated vessel systems from Elastrat Sarl (Geneva, Switzerland) or TrandoMed (Ningbo Trando 3D Medical Technology Co., Ltd., Ningbo, China), and pulsatile pumps for realistic blood flow from ViVitro Labs, Inc. (Victoria, Canada) or BDC Laboratories (Wheat Ridge, USA). A diverse array of devices should also be used during \textit{in vitro} testing to reflect those available for use in clinical practice. 

        \textit{Ex vivo} evaluation can achieve standard complexity by substituting the \textit{in vitro} testbed’s vessel system with porcine or human organs (including placenta) – although our focus in this Position Statement is on complex \textit{ex vivo} testbeds such as human cadavers, which enhance realism. \textit{In vivo} studies, typically conducted in living porcine models, inherently represent complex testbeds. While \textit{ex vivo} human cadaver studies offer superior anatomical accuracy compared to \textit{in vivo} porcine studies, certification regulations, such as U.S. Food and Drug Administration or Medical Device Regulation of the European Union requirements, must guide the pathway toward clinical studies.

        This innovation funnel methodology enables the development and evaluation of numerous potential candidates while progressively enhancing clinical validity at each stage. Examples of system complexity—simple \textit{in silico}, simple \textit{in vitro}, standard \textit{in silico}, standard \textit{in vitro}, standard \textit{ex vivo}, complex \textit{ex vivo}, and complex \textit{in vivo}—are shown in Supplementary Figure~\ref{fig:simple_in_silico} through \ref{fig:complex_in_vivo}, respectively.

    \subsection{Effectiveness Metrics}

        We identified two macro-classes of effectiveness metrics. One for \textit{in silico}, \textit{in vitro} and \textit{ex vivo} development stages which focus on technical assessment and are related to navigation (e.g., path following error). While for the \textit{in vivo} development stage there is a second macro-class of metrics which is more clinically oriented (e.g., mTICI). The Delphi consensus recommendations also show that using a range of metrics would be ideal for measuring effectiveness. However, as with testbed recommendations we acknowledge that some of these metrics may be challenging to implement and therefore we advocate a pragmatic approach in this Position Statement.
 
        The main practical challenge relates to measuring forces even though several methods for their assessment exist. Measuring contact forces \textit{in silico} during navigation is feasible with the inherent advantage that measurement does not mechanically interact with the testbed or device~\cite{Robertshaw2025, Ritter2022, Jackson2023}. Measuring contact forces on the vessel walls has also been successful \textit{in vitro}, either by using a force sensor underneath the phantom base~\cite{RafiiTari2016, Chi2020}, or by equipping phantoms with surface sensors~\cite{Fischer2023}. Base or tip forces can also be measured by either equipping the endovascular device with sensors~\cite{Deaton2023} or by extending the robot drivetrain~\cite{Tadauchi2022}. However, recording such forces may require additional equipment which could compromise MT performance. Moreover, force measurements may only be considered a useful metric if a consensus existed on what should be considered excessive. With no meaningful way of measuring applied forces in patients and relating these to clinical safety, such a consensus threshold may, for now, remain elusive. Instead of using predefined thresholds, it may be more meaningful to use expert-recorded force profiles as a reference. The difference between the force applied by the robotic system and that of an expert demonstrator — or a similar derived metric — could serve as a measure of effectiveness. However, no suitable method for this has been established in current research.
        
        Similarly, consideration must be given to ``path following error" (\textit{in silico} and \textit{in vitro}). Current metrics utilize the centerline of the vessel to determine the similarity of navigation path taken. However, this may not be ideal as the vessel centerline is not necessarily the best path to take during MT navigation. Instead, a more representative measure may be to pre-record the navigation of experts, and use the difference in path against the closest expert as an effectiveness metric.
        
        We propose that early research with a low TRL should focus on key effectiveness metrics of success rate, number of phases or steps, number of failures, number of handling errors made, path length, and procedure time, which can be measured effectively across all testbed types. As the TRL of the work increases, metrics such as fluoroscopy time and ``instrument tip speed" should be used to provide a more comprehensive analysis. For metrics such as path following error, and contact forces at the instrument base, instrument tip and vessel walls, more research is required to understand if and how these metrics impact navigation behavior and potential clinical outcomes during AI-assisted MT, in particular.

    \subsection{Clinical Perspective}

        Until now there has been no distinct pathway laid out for developing robotic MT in a way that will clearly demonstrate safety, clinical efficacy and utility (including cost effectiveness); all of which need to be demonstrated to introduce a new technology into routine clinical care.
        
        Whilst AI-assisted or autonomous robotic MT has limited direct clinical relevance at present, the pace of development is extremely rapid, so we must look ahead to what may be the relatively near future of MT practice and ensure any robotic system and/or AI-assistance can show safety, efficacy and utility against agreed standardized criteria. However, patient safety is critical and must override other considerations as regards clinical practice.
        
        This Position Statement and integral Delphi process have developed clear consensus regarding several components in the development, introduction and assessment of robotic MT. Consensus offers innovators, industry, researchers and clinicians an agreed appropriate pathway to achieve safe but rapid development, assessment and ultimately implementation of AI-assisted robotic MT into routine clinical practice. The effectiveness measures agreed upon for efficacy and safety \textit{in vivo} are well understood, validated, and widely used in MT research~\cite{Saver2016,Goyal2016}.
    
        It will be imperative going forwards that the safety metrics recommended in this Position Statement (for example, contact forces of vessel walls and instruments) are developed and independently validated. As an early priority next step, it would be vital to correlate in detail how \textit{in vitro} measurements (such as contact forces) relate to \textit{in vivo} safety and complications (for example, risk of perforation, intracranial hemorrhage, dissection, and distal embolization). This will require close collaboration and partnership between industry, academia and neurointerventional clinicians.

    \subsection{Patient Perspective}

        The patient representatives involved in this initiative have each experienced the devastating impact that a stroke can have on families. When a stroke is diagnosed, the urgency for intervention is unquestioned. However, the procedures are complex and risky for the patient, requiring specialized interventions. Robotic MT (with or without AI) has the potential to dramatically change this.
        
        We see that this Position Statement is the first step in addressing an uncertainty which is delaying progress towards the introduction of robotic MT (with or without AI). One reason for this delay is that there is no evidence or data comparing manual MT and robotic MT (with or without AI) and therefore it is unknown whether there is an advantage of one approach over another. Clinical participants were united in the view that common reference standards do not exist, nor do standard measures of effectiveness of treatments, but that their development was necessary.
        
        From the standpoint of patient representatives, there appears to be extensive agreement from the professions on why it is important that robotic MT (with or without AI) is translated. There also appears to be overall agreement on what is important in terms of which ``biological" characteristics need to be included (such as atheromatous plaques), and which robotic behaviors are important.
        
        Anyone with experience of this condition will want this vision to come to fruition in the clinic as quickly as possible. However, it is understood that this new technology will require rigorous proving through evidence - including clinical trials - to move beyond TRL3. Clearly the next steps must capitalize heavily on the consensus so far; and patients would then expect multiple studies, but, crucially, based upon this Position Statement derived from the Delphi process. In addition, but for future consideration, the patient community will have concerns over the evolution of robotic MT procedure towards a fully autonomous one using AI. The role of oversight – and what back up is considered acceptable in remote locations should complications occur, the development of trust in AI, and how consent would be given for this procedure will be suitable future topics for the START initiative to address. 

    \subsection{Health Economics}

        Health technology assessment (HTA) with health economic evaluation can be performed at any time during the development of a new technology and ideally should take an iterative approach alongside the development process~\cite{Clarkson2018}. At this early stage in the development of robotic MT (with or without AI), a developer-focused health technology assessment (dfHTA) approach is recommended~\cite{Bouttell2020}. This approach is essential for clearly identifying the clinical need and for generating robust evidence to support it.
        
        Importantly the dfHTA should be undertaken even before any AI-assisted robot has been finalized so that the developers can respond to the findings such as changing the overall design, use case, target population and/or place in the clinical pathway. The process of following a dfHTA approach will allow the technology developers to build evidence generation plans convergent on their technology development pathway. Qualitative research methods should be used to capture content from semi-structured discussions with regulators, policy decision makers, and expert opinion. For health economic evaluation to be useful at this stage in the development pathway, the focus of this work must be on reducing any uncertainty in the health economic model while robotic MT (with or without AI) is undergoing testing. This can be done by exploring studies of similar techniques if they exist (e.g., summarizing endovascular robots with AI-assistance~\cite{Crinnion2022}), or previous generations of the technology (e.g., showing neuroendovascular robots without AI-assistance~\cite{Robertshaw2023}) as well as exploring the reliability of the experts’ opinions on plausible ranges of effect sizes~\cite{Bojke2021}, gathering detailed information on who will be using the robot, and the impact that it will have on workflow within the clinical environment where it will be deployed. Other work could be to use a Markov model to simulate health outcomes and costs of the treatment over a lifetime time horizon, as has already been carried out for MT~\cite{Lobotesis2016}. Cost-based analyses would incorporate patient and social benefits from intervening earlier including more rapid discharge to home, more rapid return to work, and more long-term independence. Once metrics have been decided – using this Position Statement which has (1) determined which metrics are appropriate, and (2) articulated a developer-driven approach to deciding which of these metrics to use - then simulations can be run to understand the sensitivity of robotic MT (with or without AI), to each of these parameters, and to prioritize the collection of \textit{de novo} data where there is large uncertainty in the values.
        
        At this stage of the technology, it is recommended that developers focus on articulating and quantifying a value proposition by using qualitative research methods to capture users’ and patients’ lived experiences as well as some simple budget impact analyses on existing healthcare services. There will be some commonality between different robots able to perform MT (with or without AI), and in future work, START could enable capture of baseline data.

    \subsection{Ethical Considerations for Implementation}

        Autonomous or assisted navigation will be employed to mitigate the risks of a tele-operation by an expert, and/or to upskill local generalist operators at the patient’s local site. To optimize accuracy, AI-assistive technology will benefit from data-driven approaches with physical modeling of biology, physiology, and interventional devices – which means health data needs to be shared, and sharing should occur within an ethical framework.
        
        Patient safety and informed consent – also highlighted within the Clinical and Patient Perspective Positions above, respectively - must be given top priority in the ethical implementation of new technologies, guaranteeing that patients are completely aware of the risks and benefits. Patient's opinions should continue to be addressed with engagement potentially through surveys or representation during stakeholder meetings. It is noteworthy that START accrued experience of `Patient and Public Involvement' prior to the initial incubator day. Here, stakeholders were concerned that vessels would be damaged by guidewires due to a physical reality gap caused by robotic tele-operation. The very concept of AI assistance was therefore considered an important mitigation strategy for increasing safety during tele-operation.
        
        Ongoing observation after implementation of AI-assisted robotic MT is also necessary to monitor and resolve any unforeseen implications, such as differences in outcomes or access across various patient groups – especially those in the most rural regions where back up is limited should complications occur. After implementation there is also an ethical imperative to consider technology sustainability. Central to this is standardized billing and coding procedures that appropriately account for the intricacy of these technologies and will ensure their long-term viability and fair compensation for all medical professionals involved.
        
        In order to ensure ethical and regulatory compliance with healthcare standards, and to streamline the respective development process, our endeavor should also concentrate on establishing regulatory pathways that enable translation of AI-assisted robotic MT from early developmental stages to clinical practice. Here, regulatory bodies need to be involved early on (these are nation-specific and it is beyond our remit to detail these – but in the US, for example, this would include those bodies involved in state-based licensure to streamline dual licensing through the Interstate Medical Licensure Compact (IMLC), or cross-state hospital credentialing, or even policy reform for telehealth reciprocity and federal licensure). Before full implementation robotic AI-assisted MT, consideration should also be given to cybersecurity and the transmission of information required to enable tele-operated robotics, ensuring that all data remains secure in critical interventions.
        
        We can improve the integration of cutting-edge technologies and procedures in clinical settings by taking a comprehensive approach to these challenges, which will eventually improve access to patient care and endovascular intervention outcomes globally. We suggest convening a dedicated future incubator day and creating a follow-up Position Statement to address the broad implementation, and related ethical issues.

    \subsection{Strengths and Limitations}

        The mixed group of stakeholders from five countries and two continents, and the 100\% yield for every step in the Delphi process, suggests that the results represent a reasonable estimate of best practice for optimizing testbeds and effectiveness metrics. Our item list for the Delphi process was comprehensive, could be updated during the process, and was likely to capture the key endovascular testbed items and the key effectiveness metrics to be considered important by the panel. Nonetheless, this study has limitations. The panel was small, and unlikely to be representative of all stakeholders. For example, despite START being announced at multiple international conferences and meetings, many experts may be unaware of the initiative or may not have had time to contribute. Moreover, although the Delphi process was designed with care, responses may reflect varying interpretation of the questions. Furthermore, the deliberately narrow focus on testbeds and effectiveness metrics was at the detriment of focusing on other important aspects of AI-assisted robotic MT. However, the START initiative has only just begun, and as shown above, there are already recommendations for future developments including new Position Statements. Other avenues that would benefit the community through consensus also include procedural training. Additional Position Statements, especially of published within 12-24 months of this one, will together lead to greater translational impact.

\section{Conclusion}\label{sec5}

    Currently, AI-assisted robotic MT has limited clinical relevance. However, the pace of development of robotics and AI is so rapid that transformative solutions are likely to be imminent~\cite{Tang2020}. Thereafter the technology can be taken beyond TRL 3 and implemented once key translational milestones are passed (Supplementary Table~\ref{tab:translation_milestones}). Therefore, we must ensure that any robotic system and/or AI-assistance can show safety, efficacy and utility against agreed standardized criteria. Through a Delphi consensus, this Position Statement has established the first step in this process by establishing recommendations for testbeds and effectiveness metrics to develop and validate AI-assistive robotic MT. We have identified a practical approach to judiciously select these established recommendations on a case-by-case basis, recognizing the differing requirements for testbed realism – and therefore complexity – during AI-assistive robotic MT at different developmental stages. We endorse a similar practical approach regarding effectiveness metrics. We have also identified patient safety to be central, and yet not perceived to be a risk, to the development of this technology. One requisite patient safety task needed now is to correlate \textit{in vitro} measurements (for example, contact forces) to \textit{in vivo} complications (for example, risk of perforation, intracranial hemorrhage, dissection, and distal embolization). Health economic evaluation can also begin now, even at a low technology readiness level. A follow-up Position Statement will address implementation strategies and ethical considerations.

\section*{Statements and Declarations}

\paragraph{Contributions}

    Authors HR and TB served as overall editors. HR, TB and AG contributed to the conception and design of the Delphi. The individual sections were drafted by: Abstract (TB), Introduction (HR), Materials and Methods (HR), Results (HR), Discussion (TB), Endovascular Testbeds (LK, AG, FMU, MO, BJ, and HS), Effectiveness Metrics (HR, RC, DM, MP, and TV), Clinical Perspective (PW, VMP, and RB), Patient Perspective (PB, ATG, and NK), Health Economics (AB and RF), Ethical Considerations for Implementation (AH, JL, and MK), Strengths and Limitations (TB), and Conclusion (TB). All authors contributed to the article and approved the submitted version.

\paragraph{Funding}

    This work was supported by the MRC IAA 2021 Kings College London (MR/X502923/1) and the Wellcome EPSRC Centre for Medical Engineering at King’s College London (203148/Z/16/Z).
    
    AH is supported by grants from: Asahi, Balt, Scientia, Valley Baptist, GE Healthcare, and Viz.ai.
    
    MO would like to thank the FOD Economy in the call of 5G Pilot-Projects (MSCAA) for funding this work, and the internal KU Leuven C3 project (RoboGuide).

\paragraph{Declarations of Interests}

    AH is a consultant/speaker at: Medtronic, Microvention, Stryker, Penumbra, Cerenovus, Genentech, GE Healthcare, Scientia, Balt, Viz.ai, Insera therapeutics, Proximie, NeuroVasc, NovaSignal, Vesalio, Rapid Medical, Imperative Care, Galaxy Therapeutics, Route 92, Perfuze, CorTech, Shockwave, Toro and Xcath. AH is a Principal Investigator for: COMPLETE study – Penumbra, LVO SYNCHRONISE – Viz.ai, MARRS - Perfuze, RESCUE-ICAD - Medtronic. AH is on the Steering Committee/Publication committee member for: SELECT, DAWN, SELECT 2, EXPEDITE II, EMBOLISE, CLEAR, ENVI, DELPHI, DISTALS, Rapid Pulse. AH is a DSMB for the COMAND trial.
    
    DM consults for a variety of medical/surgical robotics companies, and is the original creator of the Mentice VIST simulator. 
    
    FMU is co-founder and shareholder of Ophthorobotics AG whose interests are unrelated to the presented work.
    
    MP, RF and MK are employed by Siemens Healthineers.
    
    RC is a consultant/speaker at: Terumo Neuro, Stryker, J\&J Neuro, Accandis, Phenox \& Penumbra. RC has a collaboration (non-commercial) with Mentice AB.
    
    RF is co-founder and shareholder of Tibaray Inc. whose interests are unrelated to the presented work.
    
    TV is co-founder and shareholder of Hypervision Surgical Ltd whose interests are unrelated to the presented work.
    
    TB has performed Consultancy \& Speakers Bureau Speakers services for Siemens Healthineers, Medtronic, and Bayer. Core laboratory for Microvention.

\bibliographystyle{unsrt} 
\bibliography{references}  

\newpage

    \begin{supptable}[hb!]
        \footnotesize
        \centering
        \caption{Benefits and risks without consensus after three rounds for robotic MT, both with and without AI (across all developmental stages of robotic MT from \textit{in silico} to \textit{in vitro}). Percentage agreement for each benefit and risk was taken after all rounds.}
        \label{tab:non_consensus_bens_risks}
        \begin{tabular}{lll}
        \hline
        \textbf{Domain} & \textbf{Benefit/Risk} & \textbf{Agreement} \\ \hline
        \multirow{6}{*}{\textbf{Benefit}} & To decrease patient radiation exposure & 68\% ($15/22$)\\
         & To decrease patient contrast exposure & 41\% ($9/22$)\\
         & \begin{tabular}[c]{@{}l@{}}To identify best devices or approach for a particular \\ patient\end{tabular} & 68\% ($15/22$)\\
         & To support training of new operators & 68\% ($15/22$)\\
         & To standardize procedures & 64\% ($14/22$)\\
         & \begin{tabular}[c]{@{}l@{}}To increase instrument dexterity and improve intuitive \\ navigation (reducing the workload of the operator)\end{tabular} & 73\% ($16/22$)\\ \hline
        \multirow{6}{*}{\textbf{Risk}} & Ethical issues & 45\% ($10/22$)\\
         & Patient safety & 64\% ($14/22$) \\
         & Lack of pre-clinical or \textit{ex vivo} ways to test the equipment & 68\% ($15/22$)\\
         & Societal concerns towards robots and/or AI & 36\% ($8/22$)\\
         & Patient reluctance to undergo robotic treatment & 14\% ($3/22$)\\
         & \begin{tabular}[c]{@{}l@{}}Adoption by a neurointerventional radiologist of the \\ technology\end{tabular} & 50\% ($11/22$)\\ \hline
        \end{tabular}
    \end{supptable}

    \begin{supptable}[hb!]
        \footnotesize
        \centering
        \caption{Non-consensual items after exploring which experimental factors were considered important for each developmental stage testbed (both with and without AI). Percentage agreement for each item was taken after all three rounds.}
        \label{tab:non_consensus_factors}
        \begin{tabular}{lll}
        \hline
        \textbf{Developmental Stage} & \textbf{Factor} & \textbf{\begin{tabular}[c]{@{}l@{}}Very Effective/\\ Somewhat Effective\end{tabular}} \\ \hline
        \multirow{1}{*}{\textbf{\textit{\textit{in silico}}}} & {Simulated respiration} & 43\% ($9/21$)\\ \hline
        \multirow{1}{*}{\textbf{\textit{\textit{in vitro}}}} & {Simulated respiration} & 35\% ($7/20$)\\ \hline
        \multirow{2}{*}{\textbf{\textit{\textit{ex vivo}}}} & {Simulated respiration} & 55\% ($11/20$)\\ 
        & {Simulated blood flow} & 20\% ($4/20$)\\ \hline
        \end{tabular}
    \end{supptable}

    \begin{supptable}[hb!]
        \footnotesize
        \centering
        \caption{Non-consensual items after exploring which effectiveness measures were important for each developmental stage (both with and without AI) as well as during \textit{in vivo} clinical assessment. Percentage agreement for each item was taken after all three rounds.}
        \label{tab:non_consensus_performance_measures}
        \begin{tabular}{lll}
        \hline
        \textbf{Developmental Stage} & \textbf{Effectiveness Measure} & \textbf{\begin{tabular}[c]{@{}l@{}}Very Effective/\\ Somewhat \\ Effective\end{tabular}} \\ \hline
        \multirow{5}{*}{\textbf{\textit{\textit{in silico}}}} & {Path length} & 76\% ($16/21$)\\
         & {Instrument tip speed} & 76\% ($16/21$)\\
         & {Instrument tip acceleration} & 67\% ($14/21$)\\
         & {Volume of contrast agent} & 48\% ($10/21$)\\
         & {\begin{tabular}[c]{@{}l@{}}Number of guidewire tip touches on the vessel wall\end{tabular}} & 75\% ($15/20$)\\ \hline
        \multirow{4}{*}{\textbf{\textit{\textit{in vitro}}}} & {Instrument tip acceleration} & 75\% ($15/20$)\\
         & {Volume of contrast agent} & 55\% ($11/20$)\\
         & {Fluoroscopy time} & 75\% ($15/20$)\\
         & {\begin{tabular}[c]{@{}l@{}}Number of guidewire tip touches on the vessel wall\end{tabular}} & 75\% ($15/20$)\\ \hline
        \multirow{6}{*}{\textbf{\textit{\textit{ex vivo}}}} & Path length & 71\% ($15/21$)\\
         & Path following error & 70\% ($14/20$)\\
         & Instrument tip speed & 65\% ($13/20$)\\
         & {Instrument tip acceleration} & 55\% ($11/20$)\\
         & {Volume of contrast agent} & 55\% ($11/20$)\\
         & {\begin{tabular}[c]{@{}l@{}}Number of guidewire tip touches on the vessel wall\end{tabular}} & 70\% ($14/20$)\\ \hline
        \end{tabular}
    \end{supptable}

    \begin{supptable}[hb!]
        \footnotesize
        \centering
        \caption{Example proposed technical, clinical, economic, and regulatory milestones for safe translation of robotic endovascular systems.}
        \label{tab:translation_milestones}
        \begin{tabular}{p{3.2cm} p{11cm}}
            \hline
            \textbf{Category} & \textbf{Key Performance Indicators (KPIs) / Milestones} \\ \hline
            
            \multirow{3}{=}{\textbf{Reproducible technical KPIs}} 
            & Device compatibility: Demonstrated use with standard devices (guide/sheath, aspiration, stent-retrievers) across multiple vendors; rapid tool-change workflow. \\
            & Operating room integration and interoperability: Compatible with biplane angiography suites; adequate cyber-security (e.g., ISO 27001/NIST), medical IT-network risk management (IEC 80001-1). \\
            & Teleoperation performance: Adequate end-to-end latency, jitter, link availability; automatic fail-safe modes and ``safe stop/withdraw'' on dropouts. \\ \hline
            
            \multirow{2}{=}{\textbf{Prospective human data}} 
            & Showing non-inferior morbidity and mortality outcomes. \\
            & Faster patient access. \\ \hline
            
            \multirow{2}{=}{\textbf{Financial viability}} 
            & Confirmed commissioned network pilots: funding/reimbursement including training of staff. \\
            & Health-economic analysis: cost-effectiveness (Incremental Cost-Effectiveness Ratio (ICER), Quality-Adjusted Life Years (QALYs)), capacity gain (cases/week), avoided transfers; demonstrate feasibility of different time models including 24/7 coverage. \\ \hline
            
            \multirow{3}{=}{\textbf{Regulatory pathway}} 
            & US: FDA Investigational Device Exemption (IDE) study and Premarket Approval (PMA). \\
            & EU/UK: UK Conformity Assessed (UKCA)/Conformité Européenne (CE) marking under EU MDR/UK MDR. \\
            & Expanding state-based and/or country-based licensure. \\ \hline
            
        \end{tabular}
    \end{supptable}

    \begin{suppfig}[hb!]
        \centering
        \includegraphics[width=10cm]{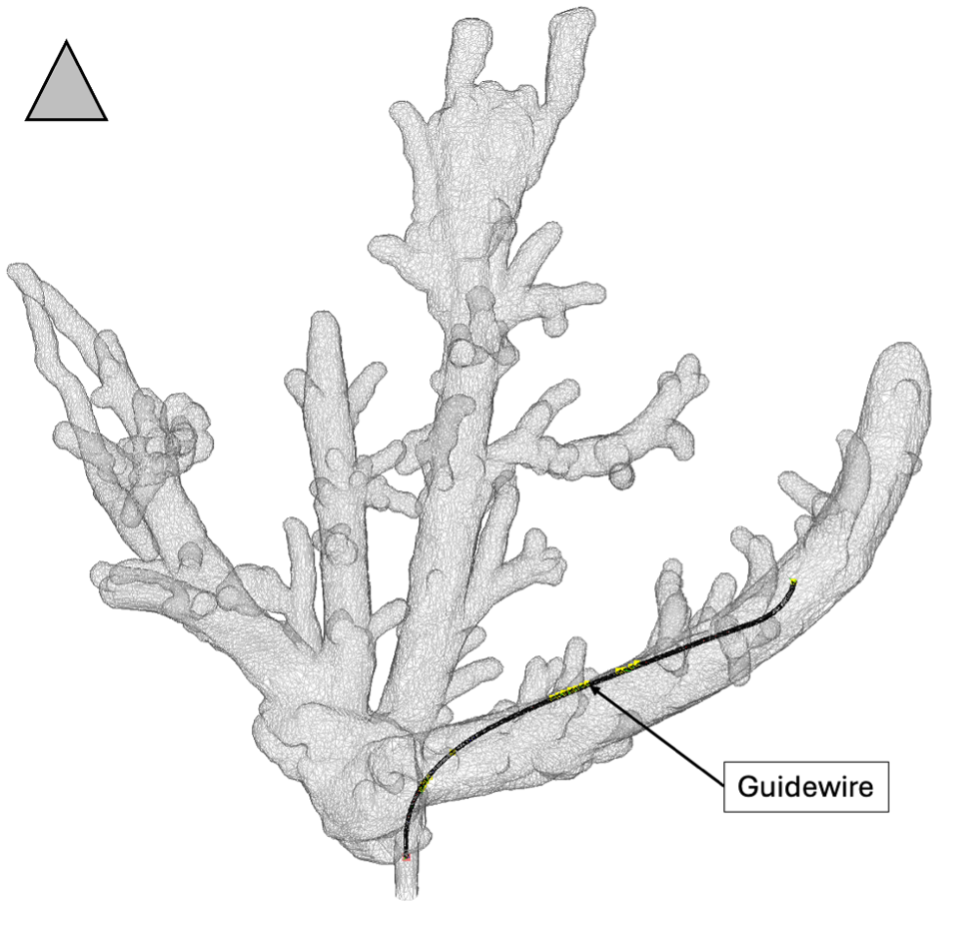}
        \caption{ Example of \textit{in silico} testbed.}
        \label{fig:simple_in_silico}
    \end{suppfig}

    \begin{suppfig}[hb!]
        \centering
        \includegraphics[width=10cm]{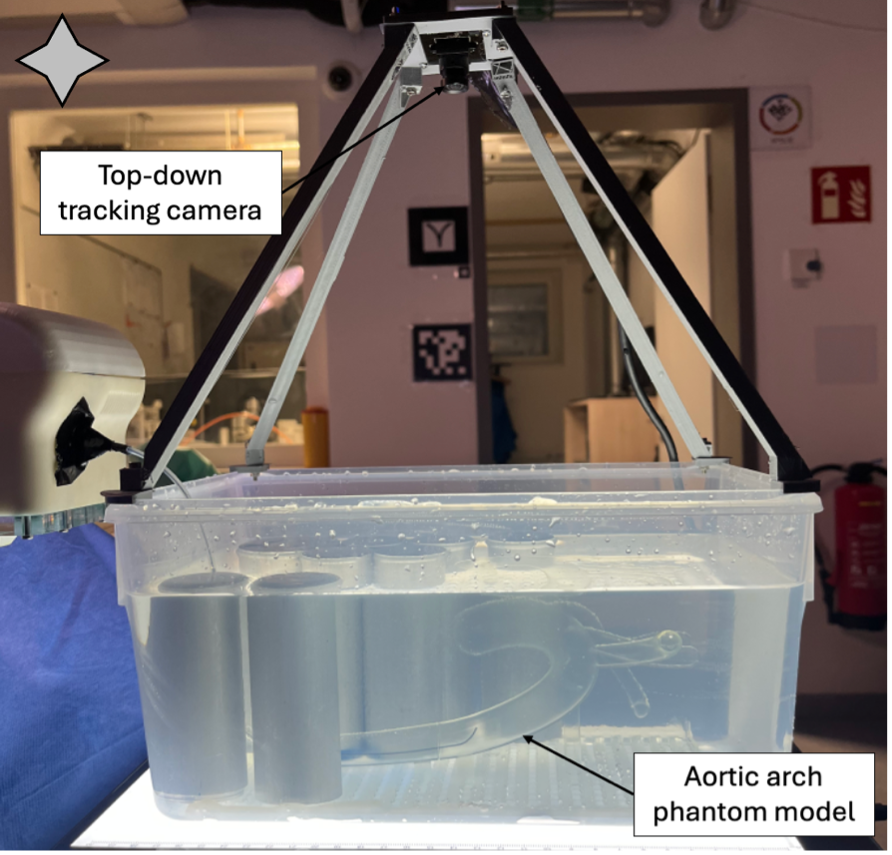}
        \caption{ Example of simple \textit{in vitro} testbed.}
        \label{fig:simple_in_vitro}
    \end{suppfig}

    \begin{suppfig}[hb!]
        \centering
        \includegraphics[width=10cm]{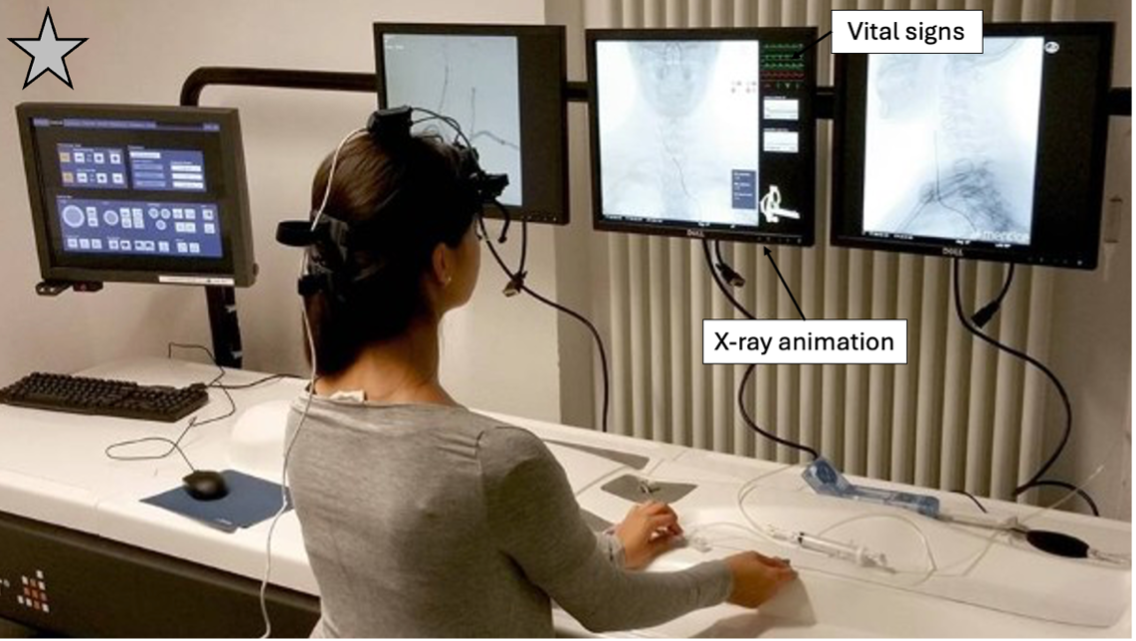}
        \caption{ Example of standard \textit{in silico} testbed. Adapted from: Kreiser et al. 2021~\cite{Kreiser2021}. Reused under the terms of the Creative Commons Attribution 4.0 International License (CC BY 4.0; http://creativecommons.org/licenses/by/4.0/). The image has been cropped and annotated by the authors.}
        \label{fig:standard_in_silico}
    \end{suppfig}

    \begin{suppfig}[hb!]
        \centering
        \includegraphics[width=10cm]{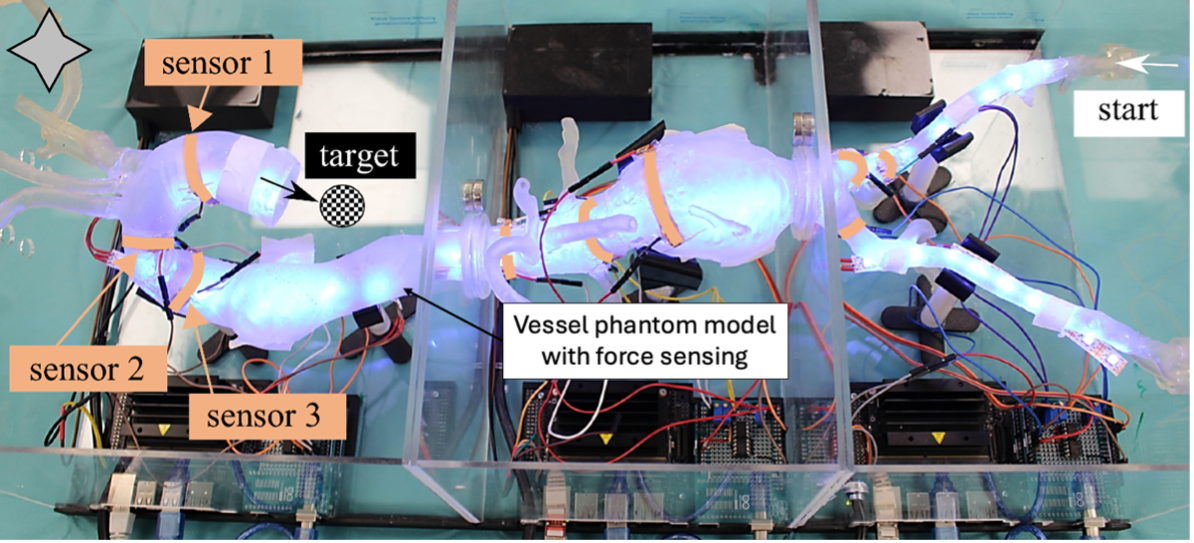}
        \caption{ Example of standard \textit{in vitro} testbed. Adapted from Fischer et al. 2023~\cite{Fischer2023}. Reused under the terms of the Creative Commons Attribution 4.0 International License (CC BY 4.0; http://creativecommons.org/licenses/by/4.0/). The image has been cropped and annotated by the authors.}
        \label{fig:standard_in_vitro}
    \end{suppfig}

    \begin{suppfig}[hb!]
        \centering
        \includegraphics[width=10cm]{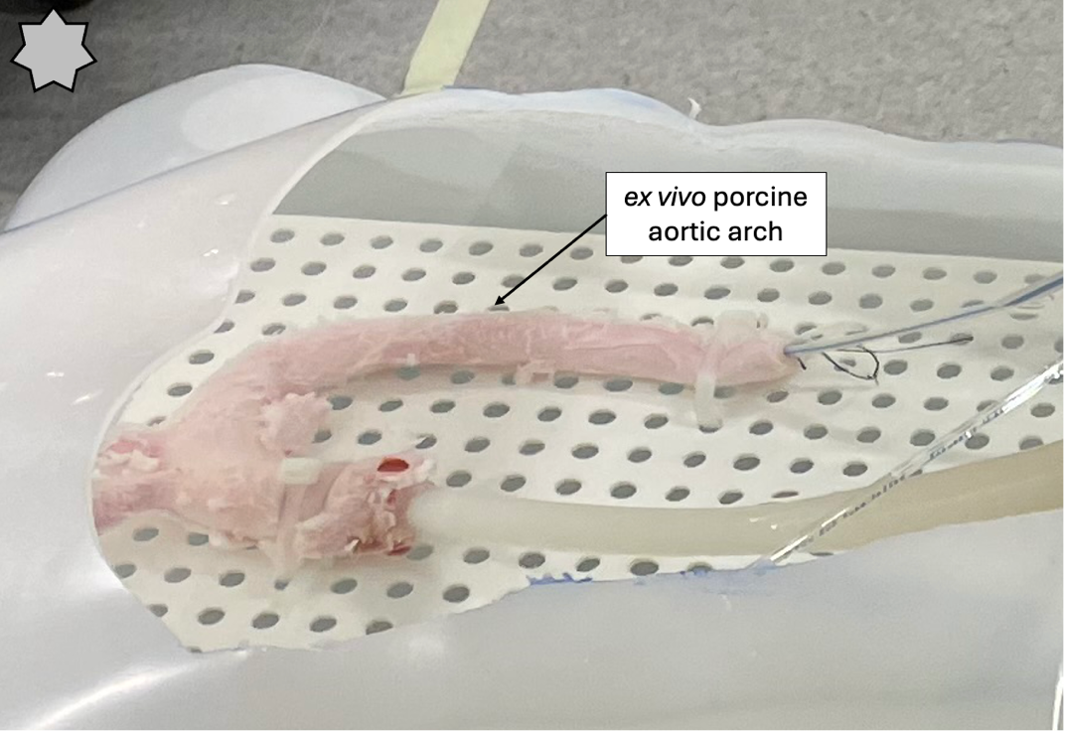}
        \caption{ Example of standard \textit{ex vivo} testbed.}
        \label{fig:standard_ex_vivo}
    \end{suppfig}

    \begin{suppfig}[hb!]
        \centering
        \includegraphics[width=10cm]{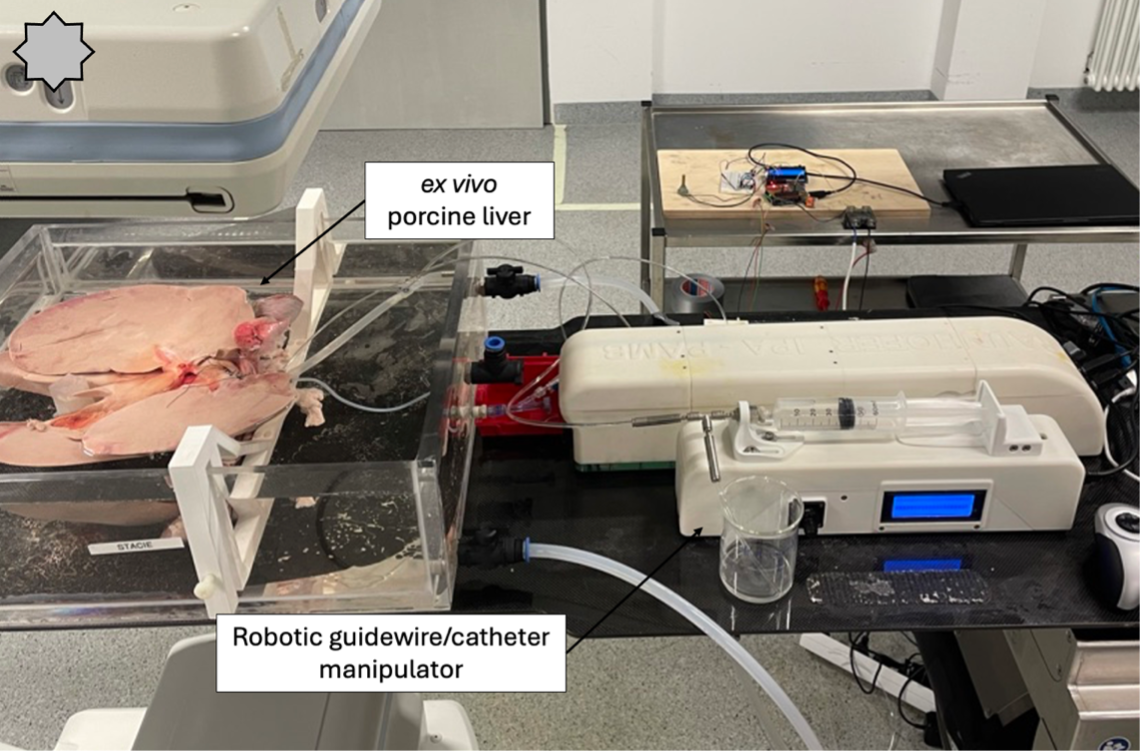}
        \caption{ Example of complex \textit{ex vivo} testbed.}
        \label{fig:complex_ex_vivo}
    \end{suppfig}

    \begin{suppfig}[hb!]
        \centering
        \includegraphics[width=10cm]{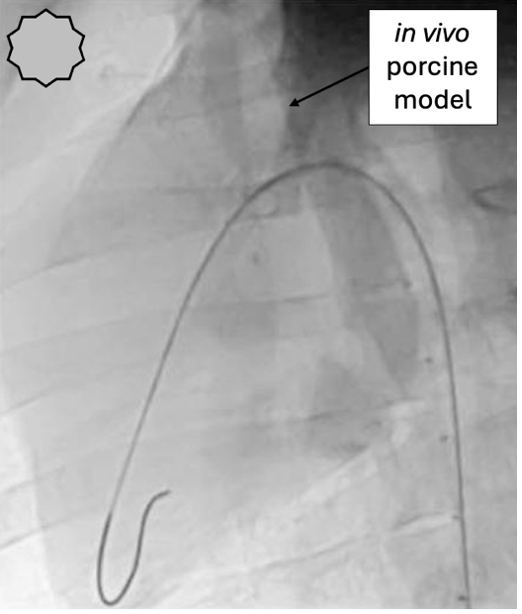}
        \caption{ Example of complex \textit{in vivo} testbed. Adapted from Peng et al. 2023~\cite{Peng2023}. Reused under the terms of the Creative Commons Attribution 4.0 International License (CC BY 4.0; http://creativecommons.org/licenses/by/4.0/). The image has been cropped and annotated by the authors.}
        \label{fig:complex_in_vivo}
    \end{suppfig}

\end{document}